\pdfoutput=1

\documentclass[11pt]{article}

\usepackage{EMNLP2022}

\usepackage{times}
\usepackage{latexsym}

\usepackage[T1]{fontenc}

\usepackage[utf8]{inputenc}

\usepackage{microtype}

\usepackage{inconsolata}

\usepackage{xspace,mfirstuc,tabulary}
\usepackage[section]{algorithm}
\usepackage[noend]{algpseudocode}
\usepackage{relsize}
\usepackage{graphicx}
\usepackage{amsmath}
\usepackage{bm}
\usepackage{amsfonts}
\usepackage{booktabs}
\usepackage{multirow}
\usepackage{mathtools}
\mathtoolsset{showonlyrefs=true}

\makeatletter
\DeclareRobustCommand\onedot{\futurelet\@let@token\@onedot}
\def\@onedot{\ifx\@let@token.\else.\null\fi\xspace}
\def\eg{\emph{e.g}\onedot} 
\def\ie{\emph{i.e}\onedot}

\makeatother

\newcommand{\Tref}[1]{Table~\ref{#1}}
\newcommand{\Eref}[1]{Equation~\eqref{#1}}
\newcommand{\Fref}[1]{Figure~\ref{#1}}
\newcommand{\Sref}[1]{Section~\ref{#1}}
\newcommand{\Appendixref}[1]{Appendix~\ref{#1}}

\algrenewcommand\algorithmicindent{0.7em}%

\title{CEFR-Based Sentence Difficulty Annotation and Assessment}

\author{
Yuki Arase$^\dagger$ \and Satoru Uchida$^\star$ \and Tomoyuki Kajiwara$^\diamond$ \\
  $^\dagger$Graduate School of Information Science and Technology, Osaka University, Japan\\
  $^\star$Faculty of Languages and Cultures, Kyushu University, Japan\\
  $^\diamond$Graduate School of Science and Engineering, Ehime University, Japan\\
  \texttt{arase@ist.osaka-u.ac.jp, uchida@flc.kyushu-u.ac.jp}\\ \texttt{kajiwara@cs.ehime-u.ac.jp} \\
}

\begin{document}
\maketitle
\begin{abstract}
Controllable text simplification is a crucial assistive technique for language learning and teaching. 
One of the primary factors hindering its advancement is the lack of a corpus annotated with sentence difficulty levels based on language ability descriptions. 
To address this problem, we created the CEFR-based Sentence Profile (CEFR-SP) corpus, containing $17$k English sentences annotated with the levels based on the Common European Framework of Reference for Languages assigned by English-education professionals. 
In addition, we propose a sentence-level assessment model to handle unbalanced level distribution because the most basic and highly proficient sentences are naturally scarce. 
In the experiments in this study, our method achieved a macro-F$1$ score of $84.5\%$ in the level assessment, thus outperforming strong baselines employed in readability assessment. 
\end{abstract}

\section{Introduction}
Controllable text simplification, first proposed by \citet{scarton-specia-2018-learning}, is the automatic rewriting of sentences to make them comprehensible to a target audience with a specific proficiency level. 
Among its primary applications are providing reading assistance to language learners and helping teachers adjust the difficulty level of their teaching materials~\cite{petersen07_slate,pellow-eskenazi-2014-open,paetzold_phdthesis}. 
The fine-grained control of output levels to match the linguistic ability of the readership is crucial for these educational applications. 

While readability assessments have been actively studied (\eg, in \citep{vajjala-thesis-2015,meng-etal-2020-text,deutsch-etal-2020-linguistic}), linking readability to language ability is difficult. 
Readability scores, such as the Flesch–Kincaid grade level \citep{fk_readability}, 
are intended for native speakers, not for language learners to whom very different considerations apply. \citet{pilan-etal-2014-rule} and \citet{ozasa2007} revealed that readability metrics designed for L$1$ do not apply to L$2$ learners.
Furthermore, readability definitions use documents rather than sentences, which are required by text simplification at the sentence-level, as their unit. 

The lack of a corpus annotated by sentence difficulty level hinders the advancement of controllable text simplification. 
Previous studies \citep{scarton-specia-2018-learning,nishihara-etal-2019-controllable,agrawal-etal-2021-non} necessarily used corpora annotated for readability rather than difficulty; furthermore, they assumed that all sentences in a document had the same readability (\ie, the document level in Newsela \citep{newsela}).

To solve these problems, we created a large-scale English corpus annotated by sentence difficulty levels based on the Common European Framework of Reference for Languages (CEFR),\footnote{\url{https://www.coe.int/en/web/common-european-framework-reference-languages}} the most widely used international standard describing learners' language ability. 
Our CEFR-based Sentence Profile (CEFR-SP) corpus adapts CEFR to sentence levels. 
A sentence is categorised as a certain level if a person with the corresponding CEFR-level can readily understand it.
CEFR-SP provides CEFR levels for $17$k sentences annotated by professionals with rich experience teaching English in higher education. 

A major challenge in sentence-level assessment is the unbalanced distribution of levels: sentences at the basic (A$1$) and highly proficient (C$2$) levels are naturally scarce. 
To handle this, we propose a sentence-level assessment model with a macro-F$1$ score of $84.5\%$. 
We designed a metric-based classification method with a simple inductive bias that avoids overfitting to majority classes \cite{vinyals-et-al-2016,protonet}. 
Our method generates embeddings representing each CEFR-level and estimates a sentence's level based on its cosine similarity to these embeddings. 
Empirical results confirm that our method effectively copes with unbalanced label distribution and outperforms the strong baselines employed in readability assessments. 

This study makes two main contributions. 
First, we present the largest corpus to date of sentences annotated according to established language ability indicators. 
Second, we propose a sentence-level assessment model to handle unbalanced label distribution. 
CEFR-SP and sentence-level assessment codes are available\footnote{The licenses of the data sources are detailed in Ethics Statement section.} for future research at \url{https://github.com/yukiar/CEFR-SP}.

\section{Related Work}
Related studies have assessed text levels on different granularity (document and sentence) and level definitions (readability/complexity and CEFR).

\subsection{Document-based Readability}
Previous studies have assessed readability and created corpora with document readability annotations. 
WeeBit \cite{weebit}, the OneStopEnglish corpus \cite{onestopenglish}, and Newsela provide manually written documents for various readability levels. 
Working with these annotated corpora, previous studies have used various linguistic and psycholinguistic features to develop models for assessing document-based readability \cite{heilman-etal-2007-combining,kate-etal-2010-learning,weebit,xia-etal-2016-text,onestopenglish}.
Neural network-based approaches have proven to be better than feature-based models \cite{azpiazu-pera-2019-multiattentive,meng-etal-2020-text,imperial-2021-bert,martinc-etal-2021-supervised}.
In particular, \citet{deutsch-etal-2020-linguistic} showed that pretrained language models outperform feature-based approaches, and the combination of linguistic features plays no role in performance gains.

\subsection{Sentence-based Readability}
Previous studies annotated sentences' {\em complexities} based on crowd workers' subjective perceptions. 
\citet{sentence_complexity_dataset} used a $5$-point scale to rate the complexity of sentences written by humans or generated by text simplification models. 
\citet{brunato-etal-2018-sentence} used a $7$-point scale for sentences extracted from the news sections of treebanks \cite{mcdonald-etal-2013-universal}. 
However, as \Sref{sec:corpus_comparison} confirms, relating complexity to language ability descriptions is challenging. 
\citet{naderi2019subjective} annotated German sentence complexity based on language learners' subjective judgements. 
In contrast, the CEFR-level of a sentence should be judged {\em objectively} based on the understanding of language learners' skills. 
Hence, we presume that a sentence CEFR-level can be judged only by language education professionals based on their teaching experience. 
For sentence-based readability assessments, previous studies regarded all sentences in a document to have the same readability \citep{collins-thompson-callan-2004-language,dellorletta-etal-2011-read,vajjala-meurers-2014-assessing,ambati-etal-2016-assessing,howcroft-demberg-2017-psycholinguistic}. 
As we show in \Sref{sec:corpus_comparison}, this assumption hardly holds. 

The {\em simplicity} of a sentence is one of the primary aspects in a text simplification evaluation, which is commonly judged by human. 
There are a few corpora annotated by the sentence simplicity for automatic quality estimation of text simplification \citep{sanja-qats-2016,alva-manchego-etal-2021-un}. 
\citet{nakamachi-etal-2020-text} applied a pretrained language model for estimating the sentence simplicity and used it to reward a reinforcement learning–based text simplification model. 
The sentence simplicity is distinctive from CEFR levels based on the established language ability descriptions.    

\subsection{CEFR-based Text Levels}
Attempts have been made to establish criteria for CEFR-level assessments. 
For example, the English Profile \cite{english_profile} and CEFR-J \citep{cefr-j} projects relate English vocabulary and grammar to CEFR levels based on learner-written' and textbook corpora. 
Tools such as Text Inspector\footnote{\url{https://textinspector.com/}} and CVLA \cite{cvla} endeavour to measure the level of English reading passages automatically. 
\citet{xia-etal-2016-text} collected reading passages from Cambridge English Exams and predicted their CEFR levels using features proposed to assess readability. 
\citet{rama-2021} demonstrated that Bidirectional Encoder Representations from Transformers (BERT) \cite{bert} consistently achieved high accuracy for multilingual CEFR-level classification.

Although these micro- (\ie, vocabulary and grammar) and macro-level (\ie, passage-level) approaches have proven useful, few attempts have been made to assign CEFR levels at the {\em sentence} level, despite its importance in learning and teaching. 
\citet{pilan-etal-2014-rule} conducted a sentence-level assessment for Swedish based on CEFR; however, they regarded document-based levels as sentence levels. 
Furthermore, their level assessment was as coarse as predicting either above B$1$ or not.

\section{CEFR-SP Corpus}
This section describes the design of the annotation procedure and discusses sentence-level profiles. 
CEFR describes language ability on a $6$-point scale: A$1$ indicates the proficiency of beginners; A$2$, B$1$, B$2$, C$1$, and C$2$ indicates mastery of a language at the basic (A), independent (B), and proficient (C) levels. 
Because CEFR is skill-based, each level is defined by `can-do’ descriptors indicating what learners can do,\footnote{\url{https://rm.coe.int/CoERMPublicCommonSearchServices/DisplayDCTMContent?documentId=090000168045bb52}} CEFR levels for sentences cannot be defined directly. 

Therefore, we used a bottom-up approach, assigning CEFR levels to sentences based on the `can-do’ descriptors of reading skills under the definition that a sentence is, for example, at A1 level if it can be readily understood by A1-level learners.
We hypothesise that with sufficient teaching experience and CEFR knowledge, it is possible to objectively determine at which level a learner can understand each sentence. 
We therefore carefully selected annotators with sufficient expertise through pilot and trial sessions. 

\subsection{Annotation Procedure}
\paragraph{Pilot Study}
A pilot study was conducted to verify the hypothesis that sufficient teaching experience and CEFR knowledge will allow an objective evaluation of sentence levels. 
We recruited participants with three levels of expertise to label $228$ sample sentences: an English-language education specialist with $12$ years of teaching experience in higher education, a graduate student majoring in English education who is familiar with CEFR, and a group of three graduate students with various majors (natural language processing and ornithology) and no prior knowledge of CEFR or English-teaching experience. 
The results showed that the second expert had a high agreement rate with the first senior expert (Pearson correlation coefficient $0.74$), whereas the members of the third group agreed less often with the senior expert (Pearson correlation coefficients: $0.45$, $0.50$, and $0.59$). 
These results confirm that annotators with considerable experience and knowledge agree on the judgement of the CEFR levels of sentences. 

\paragraph{Annotation Guidelines}
The annotators were familiarised with the annotation guidelines before beginning their work.
The guidelines described the scales and `can-do' descriptions of CEFR reading skills with example sentences of each level that were assessed by the expert. 
Importantly, the guidelines required the annotators to judge each sentence's level based on their English-teaching experience. 
Annotators were allowed to look in a dictionary to establish word levels but were instructed not to determine a sentence's level solely based on the levels of the words it contained. 

\paragraph{Annotator Selection}
For formal annotation, we recruited eight annotators with diversified English-teaching experience. 
We then conducted a trial session in which the annotators were asked to label $100$ samples extracted from the target corpora of formal annotation. 
These samples were labelled by the senior expert in the pilot study as references. 
Pearson correlation coefficients against the expert ranged from $0.59$ to $0.77$, roughly correlating with the participants' experience in English-teaching in terms of duration (years of teaching) and role (as private tutor or teacher in higher education). 
We finally selected two having high agreement rates (Pearson correlation coefficients: $0.75$ and $0.73$) and small average level-assignment differences ($0.11$ and $0.22$) compared to the expert.\footnote{CEFR levels were converted into a $6$-point scale.} 
The annotation guidelines were finalised to provide example sentences with corresponding CEFR levels on which multiple annotators had agreed in the pilot and trial sessions. 

\subsection{Sentence Selection}
Sentences were drawn from Newsela-Auto, Wiki-Auto, and the Sentence Corpus of Remedial English (SCoRE). 
Newsela-Auto and Wiki-Auto, created by \citet{jiang-etal-2020-neural}, are specifically used for text simplification.\footnote{With the plan of expanding CEFR-SP to a parallel corpus in the future, we included parallel sentences. 
Note that our data-split policy (\Sref{sec:split}) ensures that highly similar sentences do NOT appear in training and validation/test sets.} 
SCoRE \cite{score} was created for computer-assisted English learning, particularly for second language learners with lower-level proficiency. 
The sentences in SCoRE were carefully written by native English speakers, understanding the educational goals of each proficiency level; they include   
A-level sentences, which are scarce in text simplification corpora. 

The difficulty level can also be affected by external factors, such as discourse and readers' knowledge of a topic. 
For example, consider the sentence ‘The white house announced his return.' 
Though it is simple in terms of wording and grammar, understanding it requires the knowledge that `the white house' is an organisation name and the resolution of the coreference of `he (his)’ from outside the sentence. 
We consider comprehension of anaphora and cultural and factual knowledge to be different aspects of language proficiency. 
The dependence on external factors makes the sentence-level assessment ill-formed. 
To minimise the effect of outside factors, we selected {\em stand-alone} sentences for annotation, that is, sentences comprehensible independent of their surrounding context. 

Thus, we selected the first sentences in paragraphs to avoid requiring coreference resolution. 
We excluded sentences with named entities (although dates, times, country names, and numeral expressions were allowed), quotations, and brackets. 
\Appendixref{sec:sentence_selection_details} describes the complete heuristics for sentence selection. 
We conducted several rounds of manual checks by observing a few hundred samples to finalise the heuristics of the sentence selection.

After filtering, we randomly sampled $5$–-$30$ word sentences to obtain $8.5$k sentences each from Newsela-Auto and Wiki-Auto and $3.0$k sentences from SCoRE (excluding the $100$ sentences used in the trial session). 
Note that we excluded sentences from the Newsela-Auto test set so that CEFR-SP can be employed in training text simplification models in the future. 

\subsection{Sentence Profile}
\label{sec:corpus_analysis}
The two annotators independently supplied $40$k labels for the $20$k sentences. 
They assigned the same level to $37.6\%$ sentences and levels with one grade difference to $50.8\%$ sentences, which resulted in $88.4\%$ sentences with levels within one grade difference. 
Given that many sentences are likely to have intermediate levels of difficulty, we regarded both assignments as correct if they differed by only one; thus, for example, the same sentence could be labelled as both B$1$ and B$2$. 
This left us with $27,841$ labels for $17,676$ unique sentences.
\Tref{tab:examples} shows example sentences sampled from CEFR-SP. 

\begin{table}[t!]
\centering
\begin{tabular}{@{}cp{0.88\linewidth}@{}}
\toprule
A$1$ & She had a beautiful necklace around her neck.            \\
A$2$ & Some experts say the classes should be changed.          \\
B$1$ & Historically there have also been negative consequences. \\
B$2$ & Alligators are generally timid towards humans and tend to walk or swim away if one approaches.            \\
C$1$ & The metal-carbon bond in organometallic compounds is generally highly covalent. \\
C$2$ & In the past, non-photosynthetic plants were mistakenly thought to get food by breaking down organic matter in a manner similar to saprotrophic fungi.                        \\ \bottomrule
\end{tabular}%
\caption{Example sentences for each CEFR-level}
\label{tab:examples}
\end{table}

\begin{table}[t!]
\centering
\resizebox{\linewidth}{!}{%
\begin{tabular}{@{}lrrrrrr@{}}
\toprule
 &
  \multicolumn{1}{c}{\multirow{2}{*}{Num.}} &
  \multicolumn{1}{c}{\multirow{2}{*}{Length}} &
  \multicolumn{4}{c}{Lexical level} \\ \cmidrule(l){4-7} 
 &
  \multicolumn{1}{l}{} &
  \multicolumn{1}{c}{} &
  \multicolumn{1}{c}{A$1$} &
  \multicolumn{1}{c}{A$2$} &
  \multicolumn{1}{c}{B$1$} &
  \multicolumn{1}{c}{B$2$} \\ \midrule
A$1$ & $771$    & $7.7$  & $66.3$ & $15.2$ & $4.8$  & $1.3$ \\
A$2$ & $4,775$  & $10.9$ & $54.6$ & $18.2$ & $10.1$ & $3.2$ \\
B$1$ & $11,274$ & $15.2$ & $41.7$ & $20.1$ & $15.5$ & $5.9$ \\
B$2$ & $8,283$  & $18.0$ & $31.9$ & $19.1$ & $17.8$ & $7.9$ \\
C$1$ & $2,490$  & $19.0$ & $23.7$ & $16.9$ & $17.3$ & $8.5$ \\
C$2$ & $248$    & $19.2$ & $16.5$ & $15.2$ & $16.3$ & $6.8$ \\ \bottomrule
\end{tabular}%
}
\caption{Distribution of sentence lengths and lexical levels of content words (\%) in CEFR-SP}
\label{tab:lexical_profile}
\end{table}

\Tref{tab:lexical_profile} shows the number of sentences per level, average sentence length (number of words), and distribution (\%) of lexical levels computed on the content words in the $27,841$ labelled sentences.\footnote{We used Stanza \cite{qi-etal-2020-stanza} version $1.3.0$ for preprocessing.}  
We used the CEFR-J Wordlist\footnote{CEFR-J Wordlist Version $1.6$ \url{http://www.cefr-j.org/data/CEFRJ_wordlist_ver1.6.zip}}, which assigns A$1$ to B$2$ levels to pairs of lemmas and part-of-speech tags. 
This allowed us to determine word levels without word sense disambiguation.\footnote{Another possible lexicon is English Vocabulary Profile (EVP; \url{http://www.englishprofile.org/wordlists}). 
Although EVP provides C-level words, it requires word sense disambiguation to determine the level of a word, which hinders precise word-level estimation.}
The content words in sentences were matched with the CEFR-J wordlist using their lemmas and part-of-speech tags. 
The frequency of each lexical level was computed by dividing the count of words with that level by the number of all content words at each sentence-level. 
We excluded function words, assuming that they had less effect on the sentence-level. 

As expected, sentences in the A$1$ and C$2$ levels were particularly scarce. 
Sentence lengths are not proportional to CEFR levels; A level-sentences are short, whereas B-level sentences and above are similar in length. 
In contrast, the distribution of lexical levels shows a roughly positive correlation to sentence levels; A$1$-level words appear significantly more frequently in lower-level sentences, and B$1$ and B$2$ words in higher-level ones. 
A$2$-level words form an exception, appearing most frequently in the intermediate levels of A$2$ to B$2$. 

\subsection{Comparison with Existing Corpora}
\label{sec:corpus_comparison}
\begin{table}[t!]
\centering
\resizebox{\linewidth}{!}{%
\begin{tabular}{@{}crrrrrrrrr@{}}
\toprule
\multirow{2}{*}{} & \multicolumn{9}{c}{Newsela}                                                    \\ \cmidrule(l){2-10} 
                  & $2$  & $3$   & $4$    & $5$    & $6$    & $7$   & $8$   & $9$-$10$ & $11$-$12$ \\ \midrule
A$1$              & $12$ & $16$  & $35$   & $20$   & $7$    & $4$   & $3$   & $0$      & $2$       \\
A$2$              & $41$ & $148$ & $602$  & $446$  & $243$  & $172$ & $65$  & $24$     & $59$      \\
B$1$              & $30$ & $187$ & $1,155$ & $1,302$ & $1,015$ & $969$ & $475$ & $290$    & $442$     \\
B$2$              & $3$  & $37$  & $315$  & $607$  & $615$  & $709$ & $483$ & $322$    & $555$     \\
C$1$              & $0$  & $2$   & $23$   & $51$   & $59$   & $89$  & $91$  & $58$     & $176$     \\
C$2$              & $0$  & $0$   & $0$    & $1$    & $1$    & $5$   & $2$   & $3$      & $6$       \\ \bottomrule
\end{tabular}%
}
\caption{Confusion matrix between CEFR and Newsela-Auto levels: grade levels scatter across CEFR levels. 
}
\label{tab:confusion_matrix_cefr_newsela}
\end{table}

\Tref{tab:confusion_matrix_cefr_newsela} shows the confusion matrix between CEFR levels and Newsela-Auto grade levels assembled using sentences extracted from Newsela-Auto. 
Newsela assigns readability levels using Lexile and converts them into a K--$12$ grade level.\footnote{\url{https://support.newsela.com/article/grade-to-lexile-conversion/}} 
Newsela-Auto assigns the grade level of the document to all the sentences contained in it. 
The Newsela-Auto levels scatter across CEFR levels, indicating that document-based readability levels do not agree with sentence-based CEFR language ability. 

\begin{table}[t!]
\centering
\begin{tabular}{@{}crrrrr@{}}
\toprule
    & \multirow{2}{*}{Length} & \multicolumn{4}{c}{Lexical level} \\\cmidrule(l){3-6}
    &                         & A$1$    & A$2$   & B$1$   & B$2$  \\ \midrule
Lv.$1$ & $8.8$                   & $22.3$  & $10.9$ & $7.7$  & $6.3$ \\
Lv.$2$ & $13.4$                  & $17.7$  & $13.4$ & $9.9$  & $6.7$ \\
Lv.$3$ & $21.8$                  & $17.2$  & $13.7$ & $11.5$ & $7.3$ \\
Lv.$4$ & $26.8$                  & $16.5$  & $14.2$ & $12.3$ & $8.9$ \\
Lv.$5$ & $27.1$                  & $16.4$  & $9.9$  & $12.0$ & $7.3$ \\ \bottomrule
\end{tabular}%
\caption{Distribution of sentence lengths and lexical levels of content words (\%) in the sentence complexity dataset created by \citet{brunato-etal-2018-sentence}}
\label{tab:sentence_comp_data}
\end{table}

\Tref{tab:sentence_comp_data} shows the distribution of sentence lengths and lexical levels of content words (\%) in the sentence complexity corpus of \citet{brunato-etal-2018-sentence}. 
This corpus rated sentence complexity on a $7$-point scale, with $1$ indicating `very simple’ and $7$ indicating `very complex'. 
Based on this paper, we extracted sentences having degrees of agreement greater than or equal to $10$ and determined their levels as rounded means of assigned levels. 
We found that no sentences were assigned levels higher than $5$, which means this corpus lacks sentences at the most complex levels. 
In contrast, CEFR-SP provides C-level sentences, which are considered the most complex.  

The distributions in \Tref{tab:sentence_comp_data} are distinct from those in our corpus (\Tref{tab:lexical_profile}). 
Although \citet{brunato-etal-2018-sentence} reported that sentence length shows a clear correlation with complexity level, this was not true for our sentences of level B$1$ or higher. 
In \Tref{tab:sentence_comp_data}, the distribution of each lexical level across complexity levels was relatively uniform. 
In contrast, CEFR-SP showed a positive correlation between sentence and lexical levels. 
The results suggest that the standards of our CEFR-level annotations based on formal language ability descriptions were significantly different from the annotators' subjective perception of complexity.

\section{Sentence-Level Assessment}
We propose a sentence-level assessment model robust to imbalances in label distribution. 

\subsection{Problem Definition}
CEFR levels are ordinal: \eg, the B$2$ level is higher than the B$1$ level. 
It might therefore seem natural to model the level assessment as a regression problem. 
However, the gaps between the levels can be nonuniform, 
making the interpretation of regression outputs difficult; for example, we cannot decide whether an output of $0.7$ corresponds to A$1$ or A$2$ \cite{heilman-etal-2008-analysis,francois-2009-combining}. 
Therefore, we model CEFR-level assessment as a multiclass classification problem.\footnote{Moreover, a classification model was superior to a regression model in our preliminary experiments.}  

Given a training corpus with $N$ labelled samples $\{(x_0, y_0),(x_1, y_1), \cdots, (x_{N-1}, y_{N-1})\}$, where $x_i$ is a sentence and $y_i \in \{0, 1, \cdots, J-1\}$ indicates the index of the corresponding level, we train a classifier that classifies an input sentence into $J$ classes; $J=6$ in CEFR. 
For brevity, we do not distinguish between a level and its index hereafter. 

\subsection{Background: Metric-based Method}
\Tref{tab:lexical_profile} empirically shows that the distribution of sentence levels is unbalanced; the most basic and highly proficient sentences are the least common. 
An unbalanced label distribution leads to overfitting major classes and ignoring minor ones; for educational applications, such infrequent levels cannot be dismissed. 

Therefore, we propose a sentence-level assessment model that is robust against label imbalance. 
We use a metric-based approach \cite{vinyals-et-al-2016,protonet,mlman,sun-etal-2019-hierarchical} that classifies samples based on distances in a vector space, thereby avoiding overfitting by virtue of the simple inductive bias of a classifier. 
The metric-based approach has been studied for few-shot classification, where unlabelled sentences are classified by the embedding distances between labelled and unlabelled samples.    
In contrast, we explicitly learn embeddings representing CEFR levels (hereafter referred to as {\em prototypes}) and predict sentence levels using cosine similarity.  


\subsection{Metric-based Level Assessment}
\begin{figure}[t!]
\centering
\includegraphics[width=0.7\linewidth]{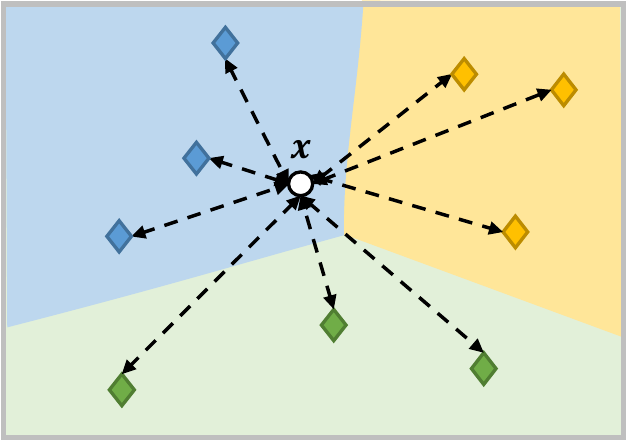}
\caption{How sentence-level is estimated by measuring similarities to level embeddings (represented by $\diamondsuit$).}
\label{fig:model}
\end{figure} 
We assume that representing a CEFR-level by a single vector may be insufficient; allowing multiple prototypes improves the expressiveness of level representation. 
We generate $K$ prototypes for each CEFR-level, \ie, $KJ$ prototypes in total, constituting a prototype matrix $\bm{C} \in \mathbb{R}^{KJ \times d}$. 
The $k$-th prototype of the $i$-th CEFR-level $\bm{c}_i^k \in \mathbb{R}^d$ has the same dimension $d$ as the sentence embedding. 
We assume that the similarity between the input sentence embedding and prototype measures the likelihood that the sentence has the corresponding label, as shown in \Fref{fig:model}. 

We employ a pretrained masked language model (MLM) to encode a sentence. 
Specifically, we encode an input sentence with $m$ tokens $x=\{w_0, w_1, \cdots, w_{m-1}\}$ using MLM to obtain the hidden outputs of each token\footnote{In practice, MLM may attach special tokens such as {\tt [CLS]} and {\tt [SEP]} to an input, which are omitted for brevity.}
\begin{equation}
    \bm{h}_0, \bm{h}_1, \cdots, \bm{h}_{m-1}=\text{MLM}(w_0,w_1,\cdots, w_{m-1}),
\end{equation}
where $\bm{h}_i \in \mathbb{R}^d$. 
We generate a sentence embedding $\bm{x} \in \mathbb{R}^d$ by mean pooling these token embeddings~\cite{reimers-gurevych-2019-sentence}: 
\begin{equation}
    \bm{x}=\text{MeanPool}(\bm{h}_0, \bm{h}_1, \cdots, \bm{h}_{m-1}). \label{eq:mean_pool}
\end{equation}

Finally, we compute the distribution $p$ over the levels for $\bm{x}$ using softmax considering similarities to the prototypes:
\begin{equation}
    p(y=j|\bm{x})=\frac{\exp(\text{CosSim}(\bm{x}, \bm{c}_j))}{\Sigma_j\exp(\text{CosSim}(\bm{x}, \bm{c}_j))},
\end{equation}
where $\text{CosSim}(\cdot, \cdot)$ calculates cosine similarity. 
When a level has multiple prototypes $K > 1$, we compute the mean of the cosine similarities:
\begin{equation}
    \text{CosSim}(\bm{x}, \bm{c}_j)=\frac{\Sigma_k\text{CosSim}(\bm{x}, \bm{c}_j^k)}{K}.
\end{equation}

\subsection{Loss Weighting}
The entire model, including MLM, is trained to minimise cross-entropy loss. 
For further alleviation of the unbalanced label distribution, loss weighting is applied according to the multinomial distribution of the level frequency \cite{conneau_and_lample_2019}.
\begin{equation}
    p_i=\frac{q_i^\alpha}{\Sigma_{i=0}^{J-1} q_i^\alpha}, \label{eq:loss_weight}
\end{equation}
where $q_i$ is the frequency of level $i$ in the training set, and $\alpha \in [0, 1]$ controls the weight strength. 
A small alpha gives large weights to infrequent labels.

\subsection{Prototype Initialisation}
The experiments established that the initialisation of prototypes affects the training stability, as the prototypes are learned from scratch. 
Therefore, the prototypes have consistent values set during initialisation to stabilise model training. 
Assuming that common characteristics of the same level of sentences are reflected in their embeddings, we use the mean of sentence embeddings in \Eref{eq:mean_pool}: $\hat{\bm{c}}_i=\text{MeanPool}(\bm{x}^i_0, \bm{x}^i_1, \cdots, \bm{x}^i_{n-1})$, where $\bm{x}^i_k$ is the $k$-th sentence embedding of level $i$ and $n$ is the number of sentences at level $i$ in the training set. 

Because a level is allowed to have multiple prototypes, an initialisation vector is generated for the $k$-th prototype at level $i$, $\hat{\bm{c}}_i^k \in \hat{\bm{C}}$, by adding Gaussian noise with mean $\mu=0$ and variance $\sigma^2$ set to $5\%$ of that computed on all elements in $\hat{\bm{c}}_0, \hat{\bm{c}}_1, \ldots, \hat{\bm{c}}_{J-1}$: 
\begin{equation}
    \hat{\bm{c}}_i^k=\hat{\bm{c}}_i+\mathcal{N}(\mu, \sigma^2).
\end{equation}
Finally, expecting these prototypes to capture the distinctive features of different levels, we orthogonalise the matrix $\hat{\bm{C}}$ and set the initial values of the prototype matrix $\bm{C}$.

\section{Evaluation}
\label{sec:cefr_eval}

\begin{table}[t!]
\centering
\resizebox{\linewidth}{!}{%
\begin{tabular}{@{}crrrrrr@{}}
\toprule
           & A$1$  & A$2$   & B$1$   & B$2$   & C$1$   & C$2$  \\ \midrule
Train      & $535$ & $3,646$ & $8,996$ & $6,636$ & $1,908$ & $100$ \\
Validation & $125$ & $568$  & $1,130$ & $821$  & $290$  & $74$  \\
Test       & $111$ & $561$  & $1,148$ & $826$  & $292$  & $74$  \\ 
\bottomrule
\end{tabular}%
}
\caption{Distribution of sentence levels in training, validation, and test sets}
\label{tab:corpus_split}
\end{table}

In this section, the proposed level assessment model is evaluated using the CEFR-SP corpus. 

\subsection{Corpus Splitting}
\label{sec:split}
We split CEFR-SP into three sets: approximately $80\%$ for training, $10\%$ for validation, and $10\%$ for the test set, as shown in \Tref{tab:corpus_split}. 
We adjusted the number of sentences for infrequent levels to preserve a reasonable number of test and validation cases.\footnote{We tentatively used the higher level among the two annotated labels for assigning a sentence into either the training, validation, or test sets.} 
In corpus splitting, we ensured that highly similar sentences did not appear in both the training and validation/test sets, as detailed in \Appendixref{sec:corpus_split_details}.

A sentence in CEFR-SP may have as many as two levels, both assignments being regarded as equally reliable. 
Therefore, the predictions during training, validating, and testing were assumed correct if they matched either of the annotated labels. 

\subsection{Evaluation Metrics}
The ability to predict {\em all} levels correctly is important for educational applications. 
As the distribution of levels was unbalanced, the models were evaluated using macro-F$1$ to penalise models that ignored minor classes. 
In addition, because CEFR levels are ordinal, the models were also evaluated using the quadratic weighted kappa \cite{weighted-kappa}.

To reduce the dependence of performance fluctuation on initialisation seeds, the experiments were conducted $12$ times with randomly selected seeds. 
We then discarded the best and worst results and reported a mean macro-F$1$ score and kappa value with a $95\%$ confidence interval. 

\begin{table*}[t!]
\centering
\resizebox{\textwidth}{!}{%
\begin{tabular}{@{}cccccccccc@{}}
\toprule
 &  & A$1$     & A$2$     & B$1$     & B$2$     & C$1$     & C$2$     & Average  & Weighted $\kappa$   \\ \midrule
\multirow{2}{*}{BoW}      & w/o lossW            & $0.0$               & $69.7$              & $76.3$             & $66.4$              & $34.7$             & $0.0$              & $41.2$             & $0.354 \pm 0.000$      \\
                          & \multicolumn{1}{l}{} & $44.2$              & $64.9$              & $73.0$             & $69.6$              & $53.8$             & $8.0$              & $52.3$             & $0.429 \pm 0.000$      \\
$k$NN                       &                      & $1.5 \pm 1.4$       & $75.2 \pm 0.7$      & $81.8 \pm 0.4$     & $66.4 \pm 0.6$      & $8.1 \pm 2.6$      & $0.0 \pm 0.0$      & $38.8 \pm 0.4$     & $0.373 \pm 0.004$      \\ \midrule
BERT                      & w/o lossW            & $12.8 \pm 9.4$      & $\bm{83.6 \pm 0.3}$ & $87.0 \pm 1.1$     & $\bm{86.7 \pm 1.2}$ & $82.9 \pm 1.5$     & $76.8 \pm 5.5$     & $71.7 \pm 1.7$     & $0.592 \pm 0.012$      \\
                          &                      & $72.7\pm 3.9$       & $82.7\pm 1.1$       & $85.5\pm 0.9$      & $86.4\pm 0.7$       & $84.9\pm 1.2$      & $83.6\pm 3.0$      & $82.5\pm 0.9$      & $0.609 \pm 0.014$      \\ \midrule
\multirow{3}{*}{Proposed} & w/o lossW            & $12.0\pm 13.4$      & $\bm{83.6\pm 0.4}$  & $\bm{87.8\pm 1.2}$ & $86.3\pm 1.4$       & $83.0\pm 0.9$      & $0.0\pm 0.0$       & $58.7\pm 1.8$      & $0.595 \pm 0.013$      \\
                          & w/o init             & $76.1 \pm 1.5$      & $80.5 \pm 1.4$      & $84.7 \pm 1.4$     & $85.7 \pm 1.3$      & $85.3 \pm 1.2$     & $88.1 \pm 2.1$     & $83.3 \pm 0.9$     & $\bm{0.628 \pm 0.017}$ \\
                          &                      & $\bm{78.0 \pm 1.3}$ & $81.4\pm 0.9$       & $86.5\pm 1.1$      & $85.9\pm 0.8$       & $\bm{85.4\pm 1.3}$ & $\bm{89.7\pm 1.6}$ & $\bm{84.5\pm 0.7}$ & $\bm{0.628 \pm 0.010}$ \\ 
\bottomrule
\end{tabular}%
}
\caption{Macro-F$1$ scores (\%) per level and quadratic weighted kappa values measured on the CEFR-SP test set; `w/o lossW' indicates a model without loss weights and `w/o init' indicates a model without initialisation using sentence embeddings. The proposed method (last row) preserves high F$1$ scores at the infrequent A$1$ and C$2$ levels and the best quadratic weighted kappa value.}
\label{tab:cefr_results}
\end{table*}

\subsection{Setting}
We used BERT-Base, cased model \cite{bert} as the pretrained MLM to encode sentences in the models that were compared.\footnote{In a preliminary experiment, we compared BERT, RoBERTa \cite{roberta}, and Sentence-BERT \cite{reimers-gurevych-2019-sentence} with different configurations and confirmed that there was no significant difference between them. Therefore, we decided to use the standard BERT-Base.} 
Specifically, we used the outputs of the $11$-th layer, which performed strongly. 
$K$, the number of prototypes of the proposed method, was set to $3$ to maximise the average macro-F$1$ of the validation set in the $1$--$10$ range. 

\paragraph{Comparison}

Because of the roughly positive correlation between the word and sentence levels (\Sref{sec:corpus_analysis}), we implemented a bag-of-words (BoW)\footnote{Word-level features performed much worse and were omitted in this experiment.} classifier using support vector machines \cite{svm} as the naive baseline. 
Moreover, as a simpler variant of metric-based classification method, we implemented a $k$-nearest neighbour ($k$NN) \cite{knn} classifier. 
We used mean-pooled token embeddings of freezed BERT as features and the cosine distance for distance computation. 
The size of $k$ was set to $6$ which marked the highest macro-F$1$ on the development set.

As the state-of-the-art baseline, we used a BERT-based classifier that outperforms conventional linguistic-feature-based classifiers in predicting passage-level readability \cite{deutsch-etal-2020-linguistic} and CEFR levels \cite{rama-2021}, as well as on simple and complex binary classification \cite{garbacea-etal-2021-explainable} of the WikiLarge corpus \cite{zhang-lapata-2017-sentence}. 
The proposed model was compared with these baselines with or without loss weighting. 

\paragraph{Ablation Study}
We investigated the effect of $K$ with an ablation study. 
We also implemented variations of the proposed method without loss weighting and initialisation based on sentence embeddings. 
The former method achieved its maximum validation macro-F$1$ score when $K=1$. 
The latter method used the same settings as the proposed method, except for prototype initialisation; it initialised the prototype embeddings using a normal distribution $\mathcal{N}(0,1)$.

\subsection{Implementation Details}
The classifier layer of the BERT baselines comprised a linear layer with weights $W \in \mathbb{R}^{d \times J}$ and a $10\%$ dropout to the input sentence embedding. 
Other conditions remained the same as those of the proposed method. 
We input a sentence embedding computed by \Eref{eq:mean_pool} and 
calculated the standard classification loss of cross-entropy. 
Loss weights were computed by \Eref{eq:loss_weight}.

All models were implemented using the PyTorch, Lightning, Transformers \cite{wolf-etal-2020-transformers}, and scikit-learn libraries.\footnote{Hyperlinks to these libraries are listed in \Appendixref{sec:links}.} 
The neural network models were trained on an NVIDIA Tesla V$100$ GPU using an AdamW~\cite{adamw} optimiser with a batch size of $128$. 
The training was stopped early, with $10$ patience epochs and a minimum delta of $1.0e-5$ based on the average macro-F$1$ score of all levels measured on the validation set. 
The loss weighting factor $\alpha$ and other hyperparameters were tuned using Optuna \cite{optuna}. 
For the proposed method and BoW and BERT baselines, $\alpha$ values were set to $0.2$, $0.3$, and $0.4$, respectively. 
The complete hyperparameter settings are described in \Appendixref{sec:hyperparameter_setting}.

\subsection{Results}
\label{sec:main_results}

\Tref{tab:cefr_results} shows the CEFR-SP test set results by means of macro-F$1$ scores (\%) per level and quadratic weighted kappa values with $95\%$ confidence intervals. 
As in previous studies, the BERT-based classifiers outperformed the BoW baselines. 
This result confirms that words and their levels, despite their importance, are not solely responsible for determining sentence levels. 
The $k$NN classifier showed higher macro-F$1$ scores than BoW without loss weighting on A$2$ and B$1$ because of the powerful BERT embeddings. 
However, it failed to identify A$1$ and C levels, which indicates the significance of addressing unbalanced label distribution.

The proposed method (last row) had the highest F$1$ scores for infrequent levels, \ie, A$1$ and C$2$, but a slightly reduced performance for the more common levels. 
We consider this acceptable, considering the method's capability to assess infrequent levels. 
Overall, the proposed method achieved the highest average macro-F$1$ score ($84.5\%$) and quadratic weighted kappa value ($0.628$). 

\begin{figure}[t!]
\centering
\includegraphics[width=0.95\linewidth]{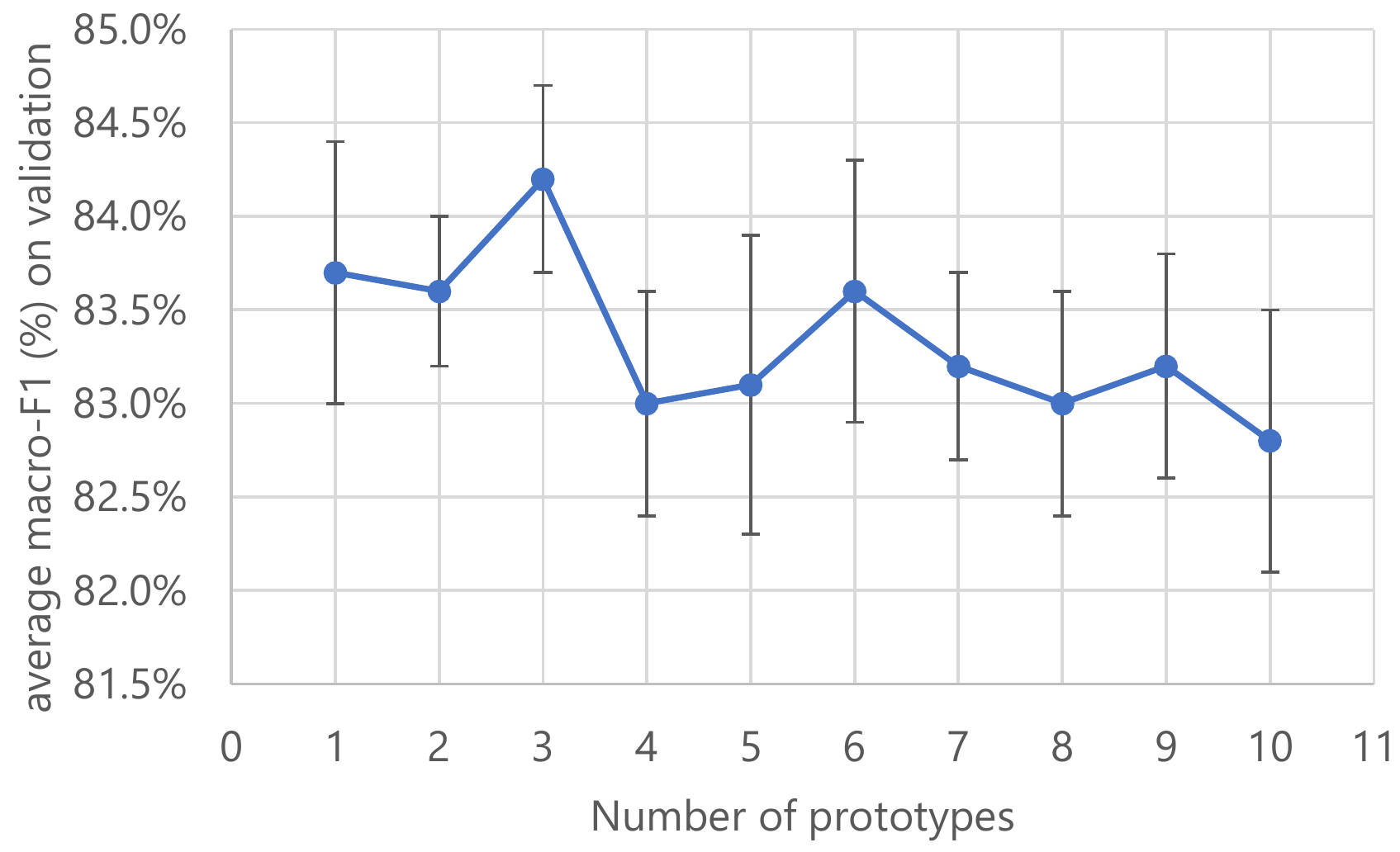}
\caption{Effects of number of prototypes: average macro-F$1$ scores (\%) measured on validation set}
\label{fig:graph_num_prototypes}
\end{figure}

\paragraph{Effects of Loss Weighting}
While loss weighting is highly effective in alleviating the effects of unbalanced label distribution on all models, it is more critical for the proposed method. 
Exclusion of loss weighting overlooks the A$1$ and C$2$ levels, as is clear from the sixth row of \Tref{tab:cefr_results}. 
Confusion matrices confirmed that A$1$ and C$2$ sentences were misclassified to their adjacent levels.

\paragraph{Effects of Initialisation}
The seventh row of \Tref{tab:cefr_results} presents the results for the proposed method without initialisation using sentence embeddings. 
This method tended to have larger confidence intervals than the proposed model. 
Moreover, we observed that it fell into an undesired solution that overlooked A$1$ and C$2$ levels depending on initialisation seeds, as reflected in lower macro-F$1$ scores. 
These results confirm that our initialisation was effective for training stabilisation.  

\paragraph{Effects of Number of Prototypes}
\label{sec:analisys_K}

\Fref{fig:graph_num_prototypes} shows the average macro-F$1$ scores with $95\%$ confidence intervals measured on the validation set when the number of prototypes in the proposed method changed from $1$ to $10$. 
The average macro-F$1$ score initially improved as the number of prototypes increased; it peaked at three, and then gradually decreased. 
This trend empirically confirms the effectiveness of multiple prototypes and shows that a relatively small number of prototypes is sufficient for CEFR-SP. 

\paragraph{Visualisation}
\begin{figure}[t!]
\centering
\includegraphics[width=0.98\linewidth]{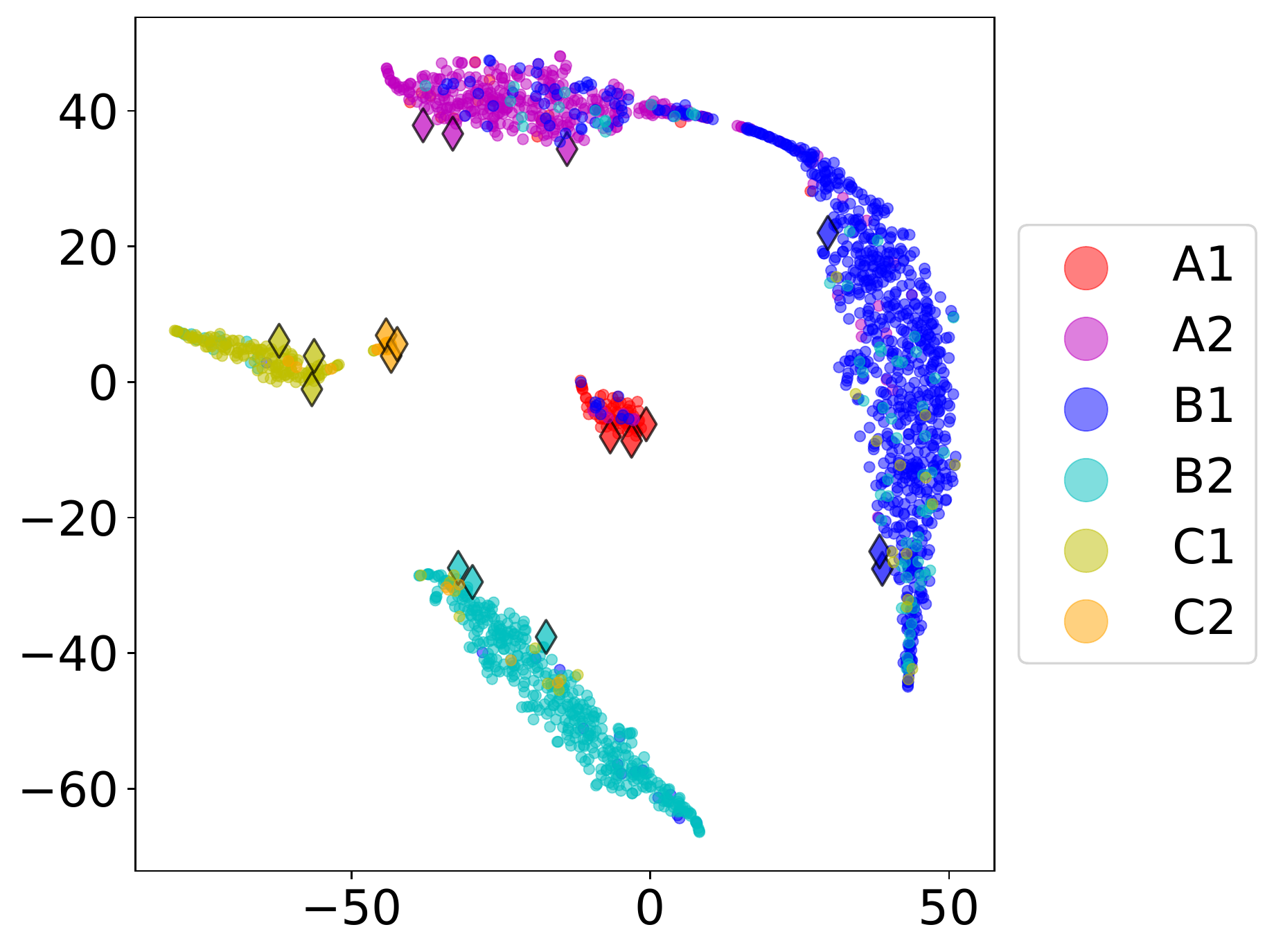}
\caption{Visualisation of prototypes (represented by $\diamondsuit$) and sentence embeddings of the proposed method.}
\label{fig:plot_proposed}
\end{figure} 

\begin{figure}[t!]
\centering
\includegraphics[width=0.98\linewidth]{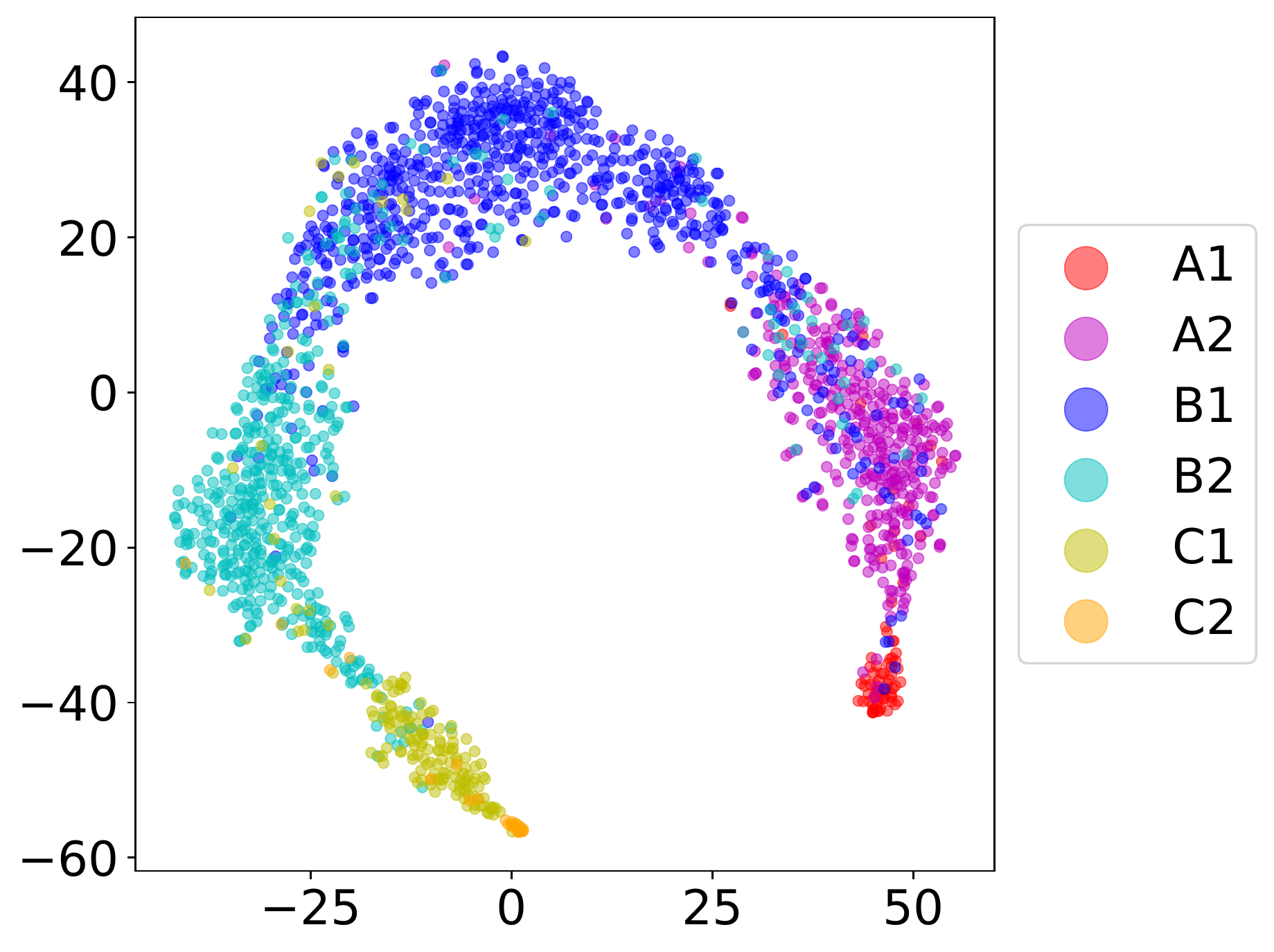}
\caption{Visualisation of sentence embeddings of BERT baseline with loss weighting.}
\label{fig:plot_baseline}
\end{figure}

\Fref{fig:plot_proposed} plots the sentence embeddings generated by the proposed method, and \Fref{fig:plot_baseline} those generated by the BERT baseline with loss weighting. 
The gold levels are colour-coded; for the proposed method, the prototypes are indicated by diamond markers. 
We used T-SNE \cite{t_sne} for visualisation, setting the perplexity to $30$ and number of iterations to $5$k to ensure convergence. 

The class boundaries were not clear in the embeddings of the baseline. 
In contrast, the embeddings of the proposed method formed clear clusters by level owing to the metric-based classification; this improved the interpretability. 
When assessing the level of a new sentence, the cosine similarity to each prototype indicates whether the assessment result is high-confidence, \ie, prototypes of a single level exhibit significantly high cosine similarity to the sentence, or ambiguous, \ie, multiple levels exhibit competitive cosine similarities.  

\section{Summary and Future Work}
In this study, we introduced CEFR-SP, the first English sentence corpus annotated with CEFR levels. 
The carefully designed annotation procedure involved recruiting experts with strong backgrounds in English education to ensure the reliability of the assigned labels. 
CEFR-SP allows the development of an automatic sentence-level assessment model. 
The proposed method can handle unbalanced level distributions using a metric-based classification. 

Our future work will involve collecting parallel sentences of CEFR-SP to make it directly applicable for training text simplification models. 
We will also develop controllable text simplification models based on reinforcement learning: the proposed level assessment model will be employed to reward the generation of lower-level sentences.

\section*{Limitations}
Because of severe space constraints, we have reported only the lexical profile of CEFR-SP. 
We will present its syntactic and psycholinguistic features and analyse it from an educational perspective in a future publication. 
Moreover, CEFR-SP is not directly applicable to train controllable text simplification models that require parallel sentences with different levels. 
Therefore, we are currently expanding CEFR-SP to make it parallel through the manual rewriting of sentences in the corpus. 
Our sentence-level assessment model helps this process. 
We can complement sentences of scarce levels by adding additional rewriting tasks. 

We suspect that the proposed method is directly applicable to other label-imbalanced classification problems. 
The empirical investigation of this is out of the scope of the present paper and is left for future work.  

\section*{Ethics Statement}
\paragraph{Ethics in Annotation Process}
The sentences in CEFR-SP were sampled from Newsela-Auto (news articles), Wiki-Auto (Wikipedia articles), and SCoRE (sentences written for an educational application of an academic project).
We believe them to be free from harmful content that insults annotators. 

We contracted with a commercial company that provides data annotation services for academia, including the management of annotators. 
We paid annotators $\$0.50$ per sentence, \ie, approximately $\$44$/h. 
This was significantly higher than the minimum wage in the place where this study was conducted, reflecting our respect for the expertise required.   

\paragraph{License Compliance}
We comply with the licenses of the original data sources of CEFR-SP. 
Specifically, we separate CEFR-SP sentences by data source and distribute them with the same license as the original datasets from which they were sampled. 
\begin{description}
\item[Wiki-Auto] CC BY-SA $3.0$
\item[SCoRE] CC BY-NC-SA $4.0$
\item[Newsela-Auto] We ask people first to obtain Newsela corpus (\url{https://newsela.com/data/}) and then contact us, following the distribution policy of Newsela-Auto.
\end{description}
For the reproducibility of the study, the training-, validation-, and test-set splits are maintained.

\section*{Acknowledgements}
We appreciate the anonymous reviewers for their insightful comments and suggestions to improve the paper. 
This work was supported by JSPS KAKENHI Grant Number JP21H03564. 

\bibliography{arase_bib}
\bibliographystyle{acl_natbib}

\appendix

\section{Details of Sentence Selection}
\label{sec:sentence_selection_details}
Dependence on external factors makes the sentence-level-assessment problem ill-formed.
This phenomenon was noticed in \citep{jacob-uitdenbogerd-2019-readability}: linguistic features that are typically well-correlated with document readability were poorly correlated with it in tweets, which inevitably depend on external factors. 
To avoid this problem, we carefully selected stand-alone sentences for annotation. 

For Wiki-Auto, we excluded the first paragraphs of an article to avoid dictionary-definition-like sentences, \eg, `X is the capital of country Y'. 
While we excluded sentences containing named entities recognised by Stanza, we allowed named entities of types of DATE, TIME, PERCENT, MONEY, QUANTITY, ORDINAL, and CARDINAL, as well as those in a list that we manually prepared containing names of well-known regions, countries, and cities (\eg, Europe, France, and Paris), and common personal names (\eg, William). 
Finally, we regularised spellings to the American forms using the localspelling library.\footnote{\url{https://github.com/fastdatascience/localspelling}} 

\section{Details of Corpus Splitting}
\label{sec:corpus_split_details}
First, we computed the cosine distances between all pairs of sentence embeddings obtained using a pretrained Sentence-BERT model~\cite{reimers-gurevych-2019-sentence}.\footnote{Specifically, we used {\tt all-mpnet-base-v2}, which had the highest performance at \url{https://www.sbert.net/docs/pretrained_models.html}.} 
Next, the average cosine distance for each sentence was calculated. 
The sentences were then allocated to the test, validation, and training sets according to the descending order of their average cosine distances. 
Thus, sentences with the least similarity to other sentences were allocated to the test and validation sets, and the rest to the training set.

\section{Hyperparameter Settings}
\label{sec:hyperparameter_setting}
For all models, the loss weighting factor $\alpha$ was searched in the range $[0.1, 1.0]$ with $0.1$ interval. 
For neural network models, the learning rate was searched in the range $[1e-5, 7e-5]$ with $1e-5$ interval. 
For the BoW baseline using support vector machines, the kernel was chosen from linear or radial basis function networks, and the regularisation parameter $\gamma$ was searched in the range $[0.01, 100]$ by loguniform sampling of $40$ points.  
\Tref{tab:hyperparam} presents the hyperparameter settings of the proposed and BERT baseline models, \Tref{tab:hyperparam_svm} those of the BoW baseline. 

\begin{table}[t!]
\centering
\resizebox{\linewidth}{!}{%
\begin{tabular}{@{}cccc@{}}
\toprule
                                  &                 & Learning Rate                  & $\alpha$ \\ \midrule
                                  & w/o lossW & $6.0e-5$ & --                    \\
\multirow{-2}{*}{BERT baseline}        &                 & $3.0e-5$                         & $0.4$                   \\
                                  & w/o lossW & $3.0e-5$                         & --                    \\
                                  & w/o init & $1.0e-5$                         & $0.2$                   \\
\multirow{-3}{*}{Proposed} &                 & $1.0e-5$                         & $0.2$                   \\ \bottomrule
\end{tabular}%
}
\caption{Hyperparameter settings of the proposed and BERT baseline models}
\label{tab:hyperparam}
\end{table}

\begin{table}[t!]
\centering
\begin{tabular}{@{}ccccc@{}}
\toprule
                     &          & Kernel & $\gamma$ & $\alpha$ \\ \midrule
\multirow{2}{*}{BoW} & w/o lossW & linear & $4.6$ & --     \\
                     &          & linear & $0.7$ & $0.3$        \\ \bottomrule
\end{tabular}%
\caption{Hyper-parameter settings of the Bag-of-Words baseline}
\label{tab:hyperparam_svm}
\end{table}

\section{Hyperlinks to Libraries}
\label{sec:links}
Here we list hyperlinks to the libraries used in implementation.
\begin{description}
\item[PyTorch] \url{https://pytorch.org/}
\item[Lightning] \url{https://www.pytorchlightning.ai/} 
\item[Transformers] \url{https://huggingface.co/docs/transformers/index}
\item[scikit-learn] \url{https://scikit-learn.org/} 
\item[Optuna] \url{https://optuna.readthedocs.io/}
\end{description}

\end{document}